\documentclass{article}
\usepackage{microtype}
\usepackage{booktabs} 
\usepackage{algorithm}
\usepackage{algpseudocode}
\usepackage{arydshln}
\usepackage{graphicx}
\usepackage{hyperref}
\usepackage[accepted]{icml2024}
\usepackage{amsmath}
\usepackage{amssymb}
\usepackage{mathtools}
\usepackage{amsthm}
\usepackage{multirow}
\usepackage{wrapfig}
\usepackage[capitalize,noabbrev]{cleveref}

\theoremstyle{plain}

\theoremstyle{definition}

\theoremstyle{remark}

\usepackage[textsize=tiny]{todonotes}

\icmltitlerunning{Active Preference Learning for Large Language Models}

\newcommand{\CodeComment}[1]{\Statex \textit{\textcolor{gray}{// #1}}}
\newcommand{\IndentCodeComment}[1]{\Statex \hspace{\algorithmicindent}\textit{\textcolor{gray}{// #1}}}
\newcommand{\E}{\mathbb{E}}

\usepackage{graphicx} % Required for including images
\usepackage{subcaption} % Required for creating subfigures
\usepackage{float}

\begin{document}

\twocolumn[
\icmltitle{Active Preference Learning for Large Language Models}

\icmlsetsymbol{equal}{*}

\begin{icmlauthorlist}
\icmlauthor{William Muldrew}{yyy}
\icmlauthor{Peter Hayes}{yyy}
\icmlauthor{Mingtian Zhang}{yyy}
\icmlauthor{David Barber}{yyy}
\end{icmlauthorlist}

\icmlaffiliation{yyy}{Centre for Artificial Intelligence,
University College London, London, UK}
\icmlcorrespondingauthor{William Muldrew}{william.muldrew.22@ucl.ac.uk}
\icmlcorrespondingauthor{Peter Hayes}{phayes@cs.ucl.ac.uk}

% You may provide any keywords that you
% find helpful for describing your paper; these are used to populate
% the "keywords" metadata in the PDF but will not be shown in the document
\icmlkeywords{LLMs, Preference Learning, Active Learning, ICML}

\vskip 0.3in
]

\printAffiliationsAndNotice{} % otherwise use the standard text.

\begin{abstract}
As large language models (LLMs) become more capable, fine-tuning techniques for aligning with human intent are increasingly important. A key consideration for aligning these models is how to most effectively use human resources, or model resources in the case where LLMs themselves are used as oracles. Reinforcement learning from Human or AI preferences (RLHF/RLAIF) is the most prominent example of such a technique, but is complex and often unstable. Direct Preference Optimization (DPO) has recently been proposed as a simpler and more stable alternative. In this work, we develop an active learning strategy for DPO to make better use of preference labels. We propose a practical acquisition function for prompt/completion pairs based on the predictive entropy of the language model and a measure of certainty of the implicit preference model optimized by DPO. We demonstrate how our approach improves both the rate of learning and final performance of fine-tuning on pairwise preference data.
\end{abstract}

\section{Introduction}
\label{submission}

Recent advancements in auto-regressive large language models (LLMs) have resulted in unprecedented capabilities in zero-shot and few-shot learning \citep{brown2020language, chowdhery2023palm}. These models are trained in an unsupervised manner using next token prediction on vast troves of mostly internet data. Their perceived capabilities and alignment with human intent are then significantly improved using various forms of fine-tuning on preference data. This fine-tuning process is a key component to producing highly capable, general purpose reasoning systems like ChatGPT.

The most prominent class of fine-tuning technique in recent times is reinforcement learning from human feedback (RLHF) \citep{ouyang2022training}. RLHF consists of a multi-stage process to adapt the pretrained autoregressive LLM $p_{\theta}(y | x)$. First, a preference data set is collected upfront. For a given prompt $x$, two completions are sampled from the model ($y_{0}, y_{1}) \sim p_{\theta}(y | x)$ and an oracle judges which they prefer. We denote $y_{w}$ as the preferred completion and $y_{l}$ as the other. Typically the oracle is a human participant, however the use of LLMs to instead provide feedback has also shown great promise \citep{bai2022constitutional}. This process is repeated over $N$ prompts resulting in the pairwise preference dataset $\mathcal{X}_{P} = \{x, y_{w}, y_{l}\}^{N}$. A reward model $r_{\phi}(x, y)$ is then trained in a supervised manner on $\mathcal{X}_{P}$. The purpose of this model is to assign a scalar score to prompt/completion pairs to measure how well they align with the oracle preferences represented by $\mathcal{X}_{P}$. Finally, a reinforcement learning (RL) algorithm such as Proximal Policy Optimization (PPO) \citep{schulman2017proximal} is used to fine-tune the parameters of the language model $\theta$ by maximising the expected reward of a different set of prompt/completion pairs $\mathcal{X}$ as measured by $r_{\phi}(x, y)$. The use of RL here circumvents the non-differentiability of sampling from $p_{\theta}(y | x)$. A downside of RLHF is its complexity; PPO introduces separate reward and value models that may be comparable in size to $p_{\theta}(y | x)$, which are typically kept in memory during training. Furthermore PPO is found to have high variance and be sensitive to choices of hyper-parameters.

Recently Direct Preference Optimization (DPO) has been proposed as a simpler and more stable alternative to RLHF \citep{rafailov2023direct}. DPO also depends on the collection of pairwise preference data, but crucially does not require first training an explicit reward model or the subsequent use of RL. Instead it relies on a straight forward binary cross entropy objective that directly increases the likelihood $y_{w}$ and decreases the likelihood of $y_{l}$. The promise of this approach is that it implicitly optimizes the same objective as RLHF, without the added complexity.

Fine-tuning state-of-the-art LLMs using both of the aforementioned methods can require highly skilled domain experts, or expensive LLMs in the case of AI feedback, to produce the required preference data. In this work, we focus on how best to utilize the available preference labelling budget, specifically when using the DPO objective to avoid the need for RL. Instead of randomly selecting a large fixed number of prompts upfront and acquiring oracle labels for a subsequent fine-tuning process, we introduce an iterative data acquisition and fine-tuning loop that we refer to as Active Preference Learning (APL). At each step, a batch of prompt/completion pairs is selected according to an acquisition function, oracle labels are acquired and then the model is improved with a cycle of fine-tuning. This loop is then repeated until some preference label budget is reached. 

We develop a simple and effective acquisition function for prompt/completion pairs that uses the predictive entropy of the latest version of the model $p_{\theta_{t}}(y | x)$ and a measure of certainty of DPO's implicit preference model. Our active sampling approach biases the fine-tuning process towards correcting data points where the models implicit preference ranking is confidently wrong; leading to better learning outcomes. We also leverage an LLM oracle to provide preference labels online and use the latest version of the fine-tuned model to generate completions at each step.

 In our experiments over multiple data sets using open source models with $\approx 1$ billion parameters, we demonstrate our approach improves the win-rate performance of the fine-tuned model by on average ~1-6\%.

\section{Direct Preference Optimization}

During the reward modelling phase in RLHF, the preference data is assumed to follow the Bradley-Terry (BT) model \citep{bradley1952rank}. The objective for training the reward model can be framed as a binary classification task with a cross entropy objective:
\begin{equation}
\mathcal{L}_{\phi}(\mathcal{X}_{P})= - \mathbb{E}_{ \mathcal{X}_{P}} [\log \sigma(r_{\phi}(x, y_{w}) - r_{\phi}(x, y_{l}))]. 
\label{eq:reward-model-objective}
\end{equation}
During the subsequent RL fine-tuning phase, the trained reward model is then used to score prompt/completion pairs to provide feedback to the language model. The aim is to maximise the following objective w.r.t $\theta$ 
\begin{equation}
\mathbb{E}_{x \sim \mathcal{X}, y \sim p_{\theta}(y | x) } [r_{\phi}(x, y)] - \beta \mathrm{KL}(p_{\theta}(y | x) || p_{\theta_{0}}(y | x)).
\label{eq:ppo-objective}
\end{equation}
The second term here regularises the fine-tuned model using the KL-divergence to stay close to the state of the LLM before fine-tuning $p_{\theta_{0}}(y | x)$. The main rationale provided for this is to prevent the model from deviating too far from the distribution on which the reward model is accurate.

In practise the following reward function is used with PPO to update $\theta$ \cite{ziegler2020finetuning, stiennon2022learning}:
\begin{equation}
r_{ppo}(x, y) = r_{\phi}(x, y) - \beta(\log p_{\theta}(y|x) - \log p_{\theta_{0}}(y | x)).
\end{equation}
DPO is derived from the optimal solution to \ref{eq:ppo-objective}; providing a maximum likelihood objective analogous to equation \ref{eq:reward-model-objective}, but parameterised by $\theta$ instead of $\phi$ \cite{rafailov2023direct};
\begin{equation}
    \mathcal{L}_{\theta}(\mathcal{X}_{P}) = - \mathbb{E}_{\mathcal{X}_{P}} 
        \left[ \log \sigma 
            \left(
               \hat{r}(x, y_{w})
                - \hat{r}(x, y_{l})         
            \right)
        \right],
\label{eq:dpo_loss}
\end{equation}
where we have the \textit{implicit reward model}
\begin{equation}
\hat{r}(x, y) = \beta \log\frac{p_\theta(y|x)}{p_{\theta_{0}}(y|x)}.
\label{eq:implicit_reward}
\end{equation}
This formulation has the distinct advantage of not requiring the explicit reward modeling step and avoids the need to perform any reinforcement learning. Furthermore, it has been shown to outperform RLHF across a range of experiments \cite{rafailov2023direct}. 

In existing work, the construction of $\mathcal{X}_{P}$ for DPO, including the preference labelling, is done upfront and stochastic gradient descent (SGD) is then used to fine-tune $\theta$ offline according to equation \ref{eq:dpo_loss} to convergence. In this work we instead assume the preference labels are not available upfront and introduce an online procedure, and that gathering said labels is expensive in time or cost as with many real world fine-tuning applications. 

\section{Active Preference Learning}
We first outline our active learning training procedure before introducing our acquisition functions for data selection. Informally, active learning is a paradigm in machine learning that aims to iteratively select the most useful datapoints during training using the current state of the model. Specifically, we are interested in the setting of pool-based active learning which involves selecting a subset of observations from a closed pool of unlabeled data \cite{ren2021survey}. 

Our APL training algorithm consists of iterations of the following scheme: randomly sample a large batch of prompts; generate pairs of completions for each prompt according to the latest version of the fine-tuned $p_{\theta_{t}}(y | x)$; rank the prompt/completion pairs according to our acquisition function; select the highest ranking subset as a batch of preference pairs for fine-tuning; query the oracle to get preference labels on this batch and, finally, fine-tune $p_{\theta_{t}}(y | x)$ using the preference labels to produce $\theta_{t+1}$. This process is repeated until some preference labelling budget has been reached.
% \ph{Might be interesting to consider non-closed pool settings where the LLM itself augments the data based on what would be useful}

This approach requires us to augment the existing DPO fine-tuning loop, which randomly samples mini-batches from a fixed preference labeled dataset, with an outer data acquisition loop. We compute the number of data acquisition steps $T$ based on an acquisition batch size $M$ and the overall labelling budget $B$. At each step we randomly sample $S$ prompts, generate completions, then score the sampled datapoints using our acquisition function, where $M < S < N$. We then select the highest ranking $M$ datapoints to add to $\mathcal{X}_{P}$ before updating $\theta_{t}$ with a round of fine-tuning. The full process in specified in algorithm \ref{alg:APO-training}.

\begin{algorithm}
\caption{Active Preference Learning Procedure}\label{alg:APO-training}
\begin{algorithmic}[1]
\footnotesize
\CodeComment{initialise dataset of prompts}
\State $\mathcal{X} \gets \{x\}^{N}$ 
\CodeComment{initialise empty preference labelled dataset}
\State $\mathcal{X}_{P} \gets \{ \ldots\}$ 
\CodeComment{compute number of acquisition steps}
\State $T \gets \lfloor\frac{B}{M}\rfloor$  
\CodeComment{initialise model weights}
\State $\theta_{t} \gets \theta_{0}$ 
\For{$t = 1 \ldots T $}
 \IndentCodeComment{randomly sample prompts} 
 \State $\mathcal{X}_{S} \coloneqq \{x\}^{S}  \sim \mathcal{X}$  
 \IndentCodeComment{generate completions} 
 \State  $\mathcal{X}_{S} \coloneqq \{y_{0}, y_{1}, x\}^{S} \gets \text{Generate}(\theta_{t}, \mathcal{X}_{S})$ 
 \IndentCodeComment{score data using acquisition function}
 \State  $\mathcal{X}_{S}\coloneqq \{s, y_{0}, y_{1}, x\}^{S} \gets \text{Score}(\theta_{t}, \mathcal{X}_{S})$ 
 \IndentCodeComment{subset to highest scoring pairs}
 \State $\mathcal{X}_{M} \coloneqq \{y_{0}, y_{1}, x\}^{M} \gets \text{Subset}(\mathcal{X}_{S})$ 
 \IndentCodeComment{get preference labels from oracle}
 \State $\mathcal{X}_{M} \coloneqq \{y_{w}, y_{l}, x\}^{M} \gets \text{Oracle}(\mathcal{X}_{M})$ 
 \IndentCodeComment{expand preference dataset}
 \State $\mathcal{X}_{P} \gets \mathcal{X}_{P}  + \mathcal{X}_{M}$  
 \IndentCodeComment{train using DPO until some stopping criteria}
 \State $\theta_{t+1} \gets \text{Finetune}(\theta_{0}, \theta_{t}, \mathcal{X}_{P}, \beta)$  
 \IndentCodeComment{evaluate model on some held out test dataset}
 \State $\text{EvaluateUsingOracle}(\theta_{t}, \theta_{0}, \mathcal{X}_{test})$ 
\EndFor
\end{algorithmic}
\end{algorithm}

Unlike typical applications of active learning in supervised learning settings, where at each acquisition step only the scoring of observations $x$ is required, we have an additional step of also generating completions for the acquired data. This is required prior to the scoring step if our choice of acquisition function needs access to completions, which we will discuss further in section \ref{sec:acquisition_functions}. 

Implementing this scheme effectively requires careful consideration of several key design choices. In the following sections we will propose a set of acquisition functions to use in step 8. Additionally, we will discuss the implementation details of the fine-tuning procedure in step 12 including how to pick the number of fine-tuning epochs. We will also cover the choice of oracle as required by steps 10 and 13. Details around settings for $S$ and $M$ will be covered in the experiments in section \ref{sec:apl_experiments}.

% , including decisions such as whether to re-initialize the parameters between data acquisition steps and determining the appropriate number of

\subsection{Acquisition functions}
\label{sec:acquisition_functions}
In selecting scoring methods (step 8 in \ref{alg:APO-training}) we aim for options that are straightforward to implement and do not require modifications to the model architectures or the fine-tuning procedure itself. This allows for a drop in addition to existing implementations. As a result, we propose using the predictive entropy of $p_{\theta_{t}}(y|x)$ as well as a measure of certainty under the Bradley-Terry preference model, which leverages the implicit reward model in DPO.

\subsubsection{Entropy of the language model}
Prior work has shown the predictive entropy (PE) to be a well calibrated measure of uncertainty in LLMs \cite{kadavath2022language}. Therefore, if used as an acquisition function it will bias the fine-tuning process towards prompts the model is more uncertain about. The model represents a conditional distribution over possible completions. The predictive entropy is defined as:
\begin{equation}
\mathcal{H}_{p_{\theta}}(y|x) = - \E_{p_{\theta}(y|x)}[\log p_{\theta}(y|x)],
\end{equation}
where this intractable integral can be approximated with Monte-Carlo samples in practise
\begin{align}
 \mathcal{H}_{p_{\theta}}(y|x) &= - \E_{p_{\theta}(y|x)}[\log p_{\theta}(y|x)] \\
 &\approx -\frac{1}{N} \sum_{n=1}^N \log p_{\theta}(y^n|x), 
\end{align}
where we calculate $\log p_{\theta}(y^n|x)$ by summing the log probability of each token in the completion.

\subsubsection{Preference model certainty}
The predictive entropy alone does not capture the extent to which the model accurately reflects oracle preferences, which is the ultimate goal of the fine-tuning process in this setting. To address this, we turn to characteristics of the Bradley-Terry model. We define a function we refer to as the certainty of the implicit preference model using $y_1, y_2 \sim p_{\theta_{t}}(y| x)$ that is maximised when the difference between the implicit rewards (see equation \ref{eq:implicit_reward}) for $y_1$ and $y_2$ is large and minimised when it's small. Specifically, during our scoring process (step 8 in algorithm \ref{alg:APO-training}) we determine the difference in our model's predicted rankings for two different completions corresponding to the same input as
\begin{equation}
\label{eq:pref-certainty}
|\hat{r}(x^i, y^i_1) - \hat{r}(x^i, y^i_2)|.
\end{equation}
We prioritize prompt/completion pairs with higher differences during the selection of data points for fine-tuning. Our hypothesis is that data points with high values provide valuable learning opportunities. Should the model's implicit preference predictions diverge from the oracle's evaluation, especially with high certainty, prioritising these discrepancies when fine-tuning can enhance model performance.

This choice is well motivated by the behaviour of the DPO training objective (equation \ref{eq:dpo_loss}). Consider the gradient update with respect to the parameters $\theta$ 
\begin{equation}
\nabla_{\theta}\mathcal{L}_{\theta} = -\beta \mathbb{E}_{\mathcal{X}_{P}}
        \left[ w
            ( \nabla_{\theta}\log p_{\theta}(y_{w}|x)
                {-} \nabla_{\theta}\log p_{\theta}(y_{l}|x)        
            )
        \right],
\end{equation}
where $w = \sigma(\hat{r}(x, y_{l}) - \hat{r}(x, y_{w}))$ weights each sample $(x, y_{w}, y_{l}) \sim \mathcal{X}_{P}$. This gradient update can be interpreted as weighting examples by how incorrectly the implicit reward model is while accounting for the strength of the KL constraint. Early in fine-tuning, when the implicit preference model is still likely to be wrong often, our proposed acquisition strategy prioritises examples that result in substantial gradient updates, which we find to accelerate learning progress and lead to an improvement in the final performance in our experiments in section \ref{sec:apl_experiments}.

\subsubsection{A hybrid approach}
In practise we can combine both entropy and preference certainty as complimentary metrics for scoring data to exploit the strengths of both. Our hypothesis is that higher entropy prompts are more likely to give incorrect predictions from the implicit preference model. In our experiments for this hybrid approach, we first select a relatively large batch of prompts and rank them by the entropy. We then take the top subset of prompts ranked by entropy and generate the required completion pairs before scoring and ranking according to preference certainty. Finally, we take the top subset of prompt/completion pairs ranked by preference certainty and add them to our preference dataset for fine-tuning. 

\subsection{Choice of oracle}
\label{sec:oracle-choice}
Algorithm \ref{alg:APO-training} requires an oracle to provide preference judgments on pairs of completions for fine-tuning (step 10) and for evaluating against a held-out test dataset (step 13). Since we aim to generate completions using the latest version of the model at each data acquisition step, using pre-labeled datasets is not feasible. Additionally, relying on human judgments is impractical due to the need for multiple experiments with different datasets, models, acquisition functions, and seeds. To address this, we turn to state-of-the-art closed-source models offered by OpenAI. The question then becomes whether these models are suitable and, if so, which model should be chosen and how should it be prompted?

We can look to recent research to answer the first question. Recent work has suggested that LLMs are superior oracles than existing metrics \cite{chen2023exploring}. Of particular relevance is the LLM as an evaluator study carried out in \cite{rafailov2023direct} for the summarization task we also use in our experiments; they provide evidence that judgements form OpenAI's GPT-4, appropriately prompted, correlate strongly with humans. Furthermore, GPT-4 and human agreement is typically similar or higher than inter-human annotator agreement on this task.

\subsubsection{Choice of prompt}
In our experiments we require two distinct oracle prompts: one for sentiment analysis and the other for summarization - see Appendix \ref{app:oracle-prompts} for details, where we've closely followed the approach outlined in \cite{rafailov2023direct}. We ask the evaluator LLM to provide a binary preference and it's rationale according to some task specific criteria included in the prompt. In order to help mitigate against any potential bias due to the ordering of results presented to model \cite{koo2023benchmarking}, we randomly change the ordering of the positive and negative completions presented to the oracle during evaluation and fine-tuning. 

\subsubsection{Choice of base model}
\begin{figure}
\vspace{-0.3cm}
\centering
\includegraphics[scale=0.5]{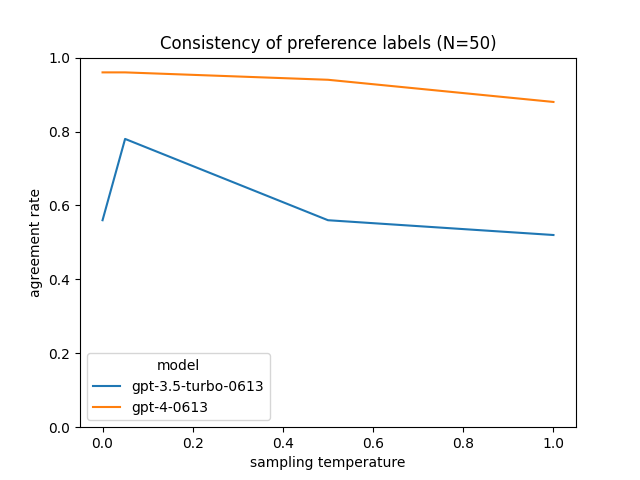}
\caption[LLM Oracle self-consistency test]{Average self-consistency of preference labels provided by GPT-3 and GPT-4 across 50 prompt completion pairs. Each model provided two preference labels for each prompt completion pair.}
\label{fig:gpt_consistency}
\end{figure}

A downside of using GPT-4 as our oracle model is the cost and high latency. A far more economical choice would be to use older versions of models such as GPT-3.5. We ran a simple analysis where we generated preference labels twice for both GPT-3.5 and GPT-4 on a set of 50 prompts and completions sampled from the fine-tuning from human preferences dataset \cite{ziegler2020finetuning}. Unfortunately, we found that GPT-4 was far more consistent ($>$90\%) than GPT-3.5-turbo (only $\sim$60\%) at a range of sampling temperatures - see figure \ref{fig:gpt_consistency}. We, therefore, chose to use GPT-4\footnote{Specifically model version \textit{gpt-4-1106-preview}} as the oracle for our experiments and adjusted our budget of evaluations appropriately. To note, our analysis assumes the same prompt for both models; we leave to future work to further prompt engineering to improve the evaluation quality and consistency of smaller, more economical models.
\subsection{Fine-tuning details}
Here we discuss in more detail the implementation details for the fine-tuning step (12) in algorithm \ref{alg:APO-training}. We adopt the most straight-forward implementation, which is to re-initialise $\theta_{t}$ to $\theta_{0}$ at each time step $t$ and fine-tune to convergence, sampling uniformly from all previously acquired preference data $\mathcal{X}_{p}$. This is consistent with previous work on deep active learning \cite{gal2017deep} and relies on the assumption that the cost (in time and/or money) of acquiring oracle labels outweighs the cost of fine-tuning again on all acquired data after each new batch of labels is acquired. The focus of our main experiments in section \ref{sec:apl_experiments} is to isolate the differences in performance caused by the different acquisition vs randomly acquiring data. In Appendix \ref{app:online}, we discuss adapting our approach for online learning and present some provisional results. 

We must also set the number of fine-tuning epochs to perform at each fine-tuning step $t$. We base this choice on an empirical analysis of the number of epochs it took on average for our choice of models to converge at different dataset sizes. Convergence was measured on the performance against a validation dataset. We analysed loss and win-rate curves for the different model and dataset combinations - see Appendix \ref{app:apl-convergence} for details.

\section{Related Work}

Our work is closely related to Direct Preference Optimization \cite{rafailov2023direct} which we leverage as our fine-tuning algorithm of choice. We augment the training process with an additional data acquisition and fine-tuning loop as outlined in algorithm \ref{alg:APO-training}. The random baseline in our experiments is equivalent to the DPO procedure. 

There are numerous recent research efforts in exploring how a more active learning setup can improve fine-tuning LLMs, but don't use DPO as a basis. The Reward rAnked FineTuning (RaFT) technique \cite{dong2023raft} introduces an online training procedure that ranks, using an oracle reward model, multiple completions for each prompt; selecting the top performers to use in a traditional supervised fine-tuning process. That is; maximising the likelihood of the best performing completions for each prompt. Once training is complete, they randomly sample a new batch of data, then re-generate completions form the latest version of the trained model and repeat the ranking/filtering and training step. Like DPO, this approach does not require the use of reinforcement learning for updating the parameters of the model. Unlike our approach, RaFT consults the oracle on every data point before filtering for the subset that will be used during training; therefore is not trying to make better use of the oracle resource. 

Another orthogonal application of active learning in the setting of improving pre-trained LLM performance is the active sampling of few shot examples for prompt stuffing \cite{margatina2023active}. In this work, the authors use acquisition functions based on different uncertainty, diversity and similarity scores of the language model across datasets of few-shot examples to determine which examples are best to reference in the prompt to improve performance. Although similar in spirit to our work, they don't consider updating the parameters of the model using preference-labelled data.

An alternative active learning approach is data pruning. In \cite{marion2023more}, pruning heuristics are applied to filter the data used in the first stage of unsupervised LLM pre-training. This leads to improved performance on downstream tasks versus the LLMs pre-trained on the full dataset. Over $50\%$ of the data can be pruned while still leading to improvements. This work does not directly consider the impact of such pruning techniques for the preference fine-tuning stage, but some of their perplexity based heuristics could represent viable alternatives to our acquisition strategies.

Finally, a research theme adjacent to active learning that can also reduce the amount of preference labels required is that of self-play fine-tuning \cite{chen2024self, yuan2024self}. These works focus on how to bootstrap $p_{\theta_{t}}(y | x)$  during fine-tuning to provide preference labels, or to act as a reward model, as opposed to trying to make better use of oracle resources. This in principle could be combined with our active preference learning approach and so we consider it complimentary.

\section{Experiments}
\label{sec:apl_experiments}

The focus of our experiments is to determine if more active sampling during the fine-tuning process can bring us gains in data efficiency when dealing with limited labelling budgets; in terms of the rate of learning and the final performance achieved. We compare four different acquisition configurations: random, entropy, certainty and entropy + certainty (as discussed in section \ref{sec:acquisition_functions}). We evaluate across two different open source large language models and two different datasets used in recent related work. We also gather some qualitative findings about the characteristics of the datapoints being acquired under the different schemes, which we discuss further in \ref{sec:qualitative-results}.

\begin{figure*}[ht]
    \centering
    \begin{subfigure}[b]{0.46\textwidth}
        \centering
        \includegraphics[width=\textwidth]{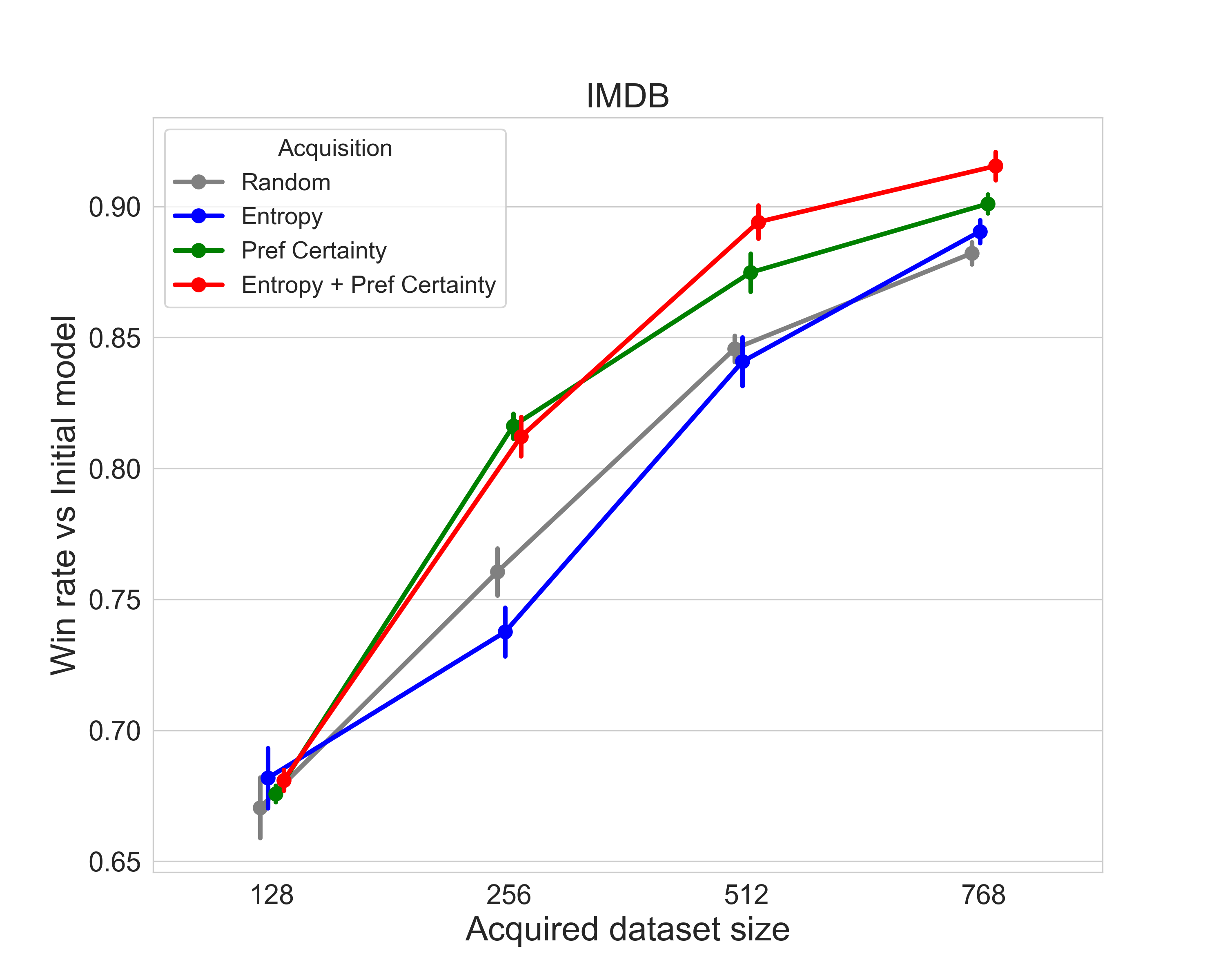}
        \caption{IMDB Win-Rate}
        \label{fig:imdb_winrate}
    \end{subfigure}
    \begin{subfigure}[b]{0.46\textwidth}
        \centering
        \includegraphics[width=\textwidth]{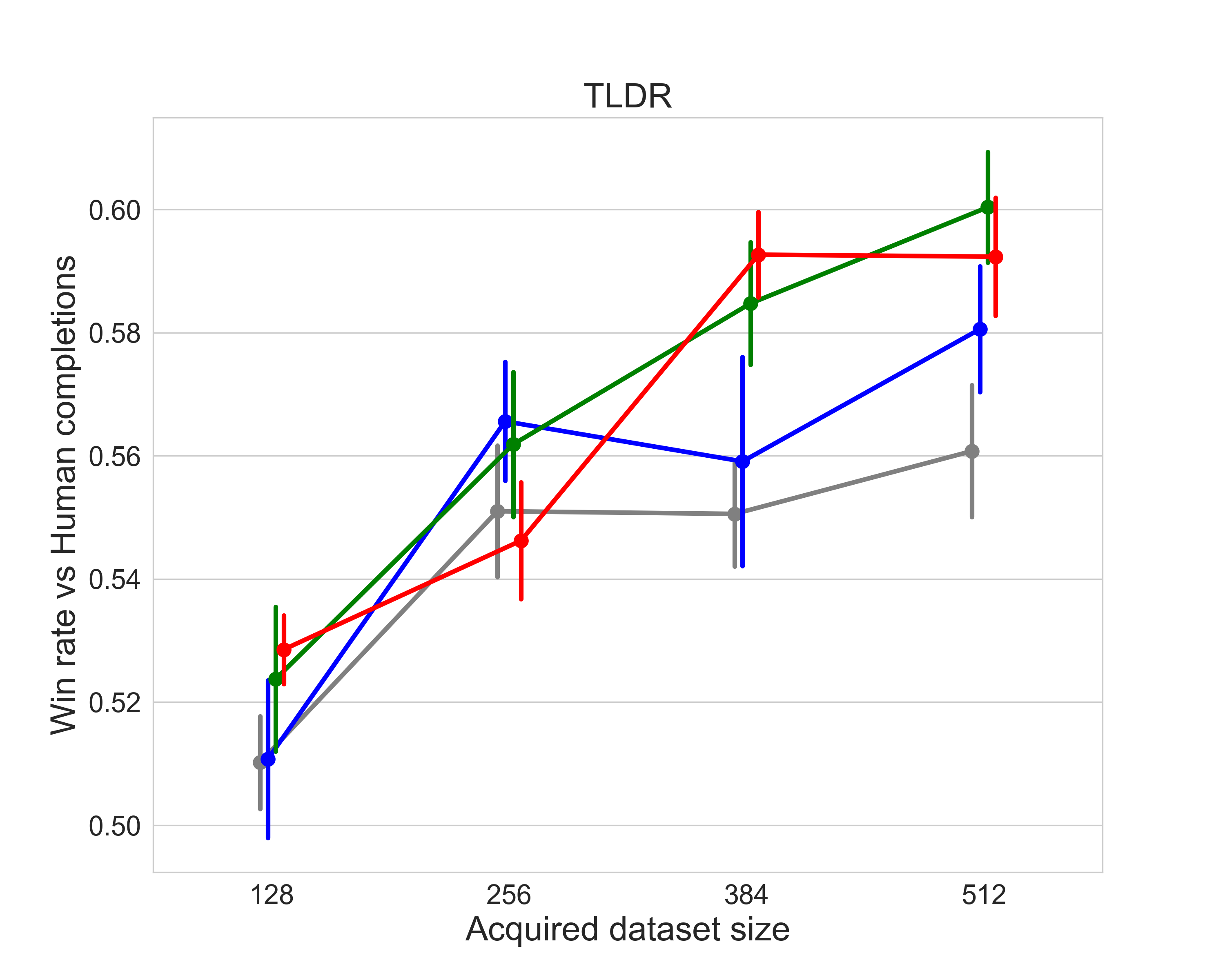}
        \caption{TLDR Win-Rate}
        \label{fig:tldr_winrate}
    \end{subfigure}
    \caption[Finetuning win-rates results at each acquisition step]{Win-rate at evaluation waypoints. (a) IMDB is win-rate vs the initial model.(b) TLDR is win-rate vs human provided summaries on the test prompts (b). The x-axis is the size of the acquired dataset used for fine-tuning at the point of evaluation. Each model and dataset pair was trained with 9 random seeds and we plot means with standard errors. Preference certainty and entropy + preference certainty  outperform the random baseline.}
    \label{fig:apl-results}
\end{figure*}

\subsection{Datasets}
In line with recent work \cite{ziegler2020finetuning, rafailov2023direct} we focus on two distinct datasets for our experiments; IMDB and TLDR. IMDB is a dataset of movie reviews where the task is to complete a positive review given the start of a review. TLDR, a more difficult task, is a dataset of Reddit posts where the task is to provide a summary of the post. Table \ref{exp:models-summary} provides a summary of the dataset details. TLDR also provides human-provided completions that can be used for evaluation.  We provide further details on dataset pre-processing in Appendix \ref{app:data-pre-processing}.

\begin{table*}[ht]
\small
\centering
\caption{\textbf{Preference learning experiments:} dataset and model details}
\label{exp:models-summary}
\begin{tabular}{p{3cm}p{3cm}p{3cm}p{4.5cm}}
\toprule
 & \textbf{IMDB} & \textbf{TLDR} & \textbf{Comment} \\
\midrule
\textbf{Task} & Complete reviews according to preference & Generate summaries according to preferences &  \\
\hdashline[.4pt/1pt]
\textbf{Train size} & 24,895 & 20,567 & Pre-processed - see \ref{app:data-pre-processing} \\
\hdashline[.4pt/1pt]
\textbf{Test size} & 24,872 & 1,159 & Pre-processed - see \ref{app:data-pre-processing}\\
\hdashline[.4pt/1pt]
\textbf{Model used} & Pre-trained GPT-2 \cite{vaswani2017attention} & Pre-trained Pythia \cite{biderman2023pythia} &  \\
\hdashline[.4pt/1pt]
\textbf{Parameter size} & 774M & 805M & \\
\hdashline[.4pt/1pt]
\textbf{Optimizer} & ADAM  lr: 1e-06 & ADAM lr:  1e-06 & As per [v1] arxiv version of DPO paper \cite{rafailov2023direct} \\
\hdashline[.4pt/1pt]
\textbf{Finetuning Epochs} & 50  & 70 & See Appendix \ref{app:apl-convergence}\\
\hdashline[.4pt/1pt]
\textbf{Mini-batch size} & 64  & 64 & For fine-tuning \\ 
\hdashline[.4pt/1pt]
\textbf{Prompt batch size (S)} & 4000  & 2048 & \\ 
\hdashline[.4pt/1pt]
\textbf{Acquisition batch size (M)} & 128  & 128 & Top M out of S examples \\ 
\hdashline[.4pt/1pt]
$\beta$ \textbf{for KL term} & 0.2  & 0.2 &  Chosen from early experiments\\
\bottomrule
\end{tabular}
\end{table*}

\subsection{Models}
\label{apl:exp-models}

For both IMDB and TLDR we use relatively large transformer based architectures. See table \ref{exp:models-summary} for a summary of the models and main hyper-parameters used in both cases. For IMDB, the GPT-2 base transformer model provided by Hugging Face\footnote{Downloaded pre-trained base model from \small\url{https://huggingface.co/edbeeching/gpt2-large-imdb}} was pre-trained on the WebText corpus \cite{radford2019language} and has 12 layers with 768 dimensions, with 12 attention heads. It was also further trained in an unsupervised way on the full IMDB dataset. 
For TLDR, we use the Pythia\footnote{Pre-trained base model from \footnotesize\url{https://huggingface.co/pvduy/pythia-1B-sft-summarize-tldr}} class of transformer model \cite{biderman2023pythia} that has an architecture similar to GPT-3, with 805M parameters, 16 layers with 2048 dimensions and 8 attention heads. We ran our fine-tuning on single 40GB RAM A100 and 48GB 600 ADAs GPUs throughout our experiments.

\subsection{Completion sampling}
We leverage temperature-scaled sampling that adjusts the probability distribution over the next token by scaling the logits before applying the softmax function. A temperature parameter $T$ controls the degree of scaling. A low temperature $T < 1$ sharpens the distribution, making the model more confident and conservative in its predictions, often leading to less diverse outputs. A high temperature $T > 1$ flattens the distribution, increasing diversity in the output by making less probable tokens more likely to be chosen. A temperature of zero $T = 0$ effectively turns the sampling into greedy decoding. In our experiments we use $T=0.7$ for $p_{\theta}(y| x)$ during training, $T=0.25$ during testing (to encourage lower variance) and $T=0.05$ for the GPT-4 oracle to promote deterministic oracle judgements. 

\subsection{Acquisition sampling}
Given we follow a pool-based active learning approach we assume access to an abundant supply of prompts to choose from during fine-tuning. In practise we have two steps to consider for filtering the data - after the initial selection of prompts (step 6 in algorithm \ref{alg:APO-training}) and after completions have been generated (step 7). In the latter case, more information is available, but require potentially expensive completions. 
    
In our experiments we first randomly sample $S=4000$ for IMDB and $S=2048$ for TLDR for our entropy only and preference certainty only acquisition runs. When doing entropy + preference certainty, we first randomly sample $J \times S$ prompts, rank them by entropy and take the top $S$ prompts to generate completions before further scoring and ranking by preference certainty. We use $J=8$ for IMDB and $J=4$ for TLDR. We use $N=8$ samples when approximating the entropy. For all experiments we set the final acquisition batch size to $M=128$.

\subsection{Evaluation}
We use GPT-4 as the oracle for providing labels and evaluating the test data. Details of the prompts are provided in Appendix \ref{app:oracle-prompts}. Our prompts specify a task-specific preference but also consider grammatical correctness and consistency. Our evaluation approach on held-out test prompts uses head-to-head win-rate comparisons versus completions sampled from the pre-trained model from the start of training $p_{\theta_{0}}(y | x)$ for IMDB.  For TLDR, we replaced the pre-trained model completions with the human-provided completions that were available on the hold-out test data. Due to the significant cost of using GPT-4 as the oracle for evaluation, we don't evaluate after every single data acquisition step. Each evaluation is done against 1024 test prompts.

\subsection{Results}

We run our active learning procedure (algorithm \ref{alg:APO-training}) to fine-tune the models discussed in the previous section against IMDB and TLDR. The overall data acquisition, fine-tuning and evaluation processes are repeated for 9 different random seeds. Figure \ref{fig:apl-results} and table \ref{exp:results-summary} contain the detailed win-rate results of each configuration. The cost associated to evaluating using GPT-4 limited the number of data acquisition steps we could practically carry out, therefore we focused on doing more seeds on fewer numbers of data acquisition steps to aid in drawing conclusions.

Overall we find that our certainty acquisition function outperforms random and entropy, improving win-rate performance by between 1-6$\%$ on average. This provides evidence in favour of our hypothesis discussed in \ref{sec:acquisition_functions} that prompts with higher differences in the implicit rewards corresponding to their completions provide valuable learning opportunities. We find that combining preference certainty with entropy gives a small improvement for the larger acquisition batch sizes (512, 768) on IMDB, but this result is not consistent across both datasets. Given these results and the additional complexity due to the Monte Carlo estimation of the entropy, we recommend the preference certainty acquisition as a simple acquisition strategy to use in practise.

For the first fine-tuning step ($M=128$), there is no discernible difference between the strategies. This makes sense when using the preference certainty acquisition because the initial pre-trained model is used to rank the data and it doesn't yet know anything about the oracle's preferences. In Appendix \ref{app:apo-samples} we provide examples of typical prompt and completion pairs, alongside the oracle preference and rationale provided by our GPT-4 oracle, before and after the fine-tuning process.

\begin{table*}[ht]
\small
\centering
\caption{\textbf{Active preference learning results:} the mean to 2 d.p. and standard errors to 3 d.p. of the win-rates. For IMDB, we calculate the win-rate vs the completions generated by the initial pre-trained model. For TLDR we calculate the win-rate vs the human completions available on the test set. The size column represents the size of the acquired dataset used for fine-tuning at the point of evaluation.}
\label{exp:results-summary}
\begin{tabular}{p{1.5cm}p{1.5cm}p{2.2cm}p{2.2cm}p{2.2cm}p{2.2cm}}
\toprule
\textbf{Dataset} & \textbf{Size} & \textbf{Random} & \textbf{Entropy} & \textbf{Pref certainty} & \textbf{Pref + Ent} \\
\midrule
\multirow{4}{*}{\textbf{IMDB}}
& \textbf{128} & $0.67 \pm 0.012$ & $0.68 \pm 0.011$ & $\textbf{0.68} \pm 0.003$ & $\textbf{0.68} \pm 0.004$ \\
& \textbf{256} & $0.76 \pm 0.008$ & $0.74 \pm 0.009$ & $\textbf{0.82} \pm 0.005$ & $0.81 \pm 0.007$ \\
& \textbf{512} & $0.84 \pm 0.004$ & $0.84 \pm 0.009$ & $0.87 \pm 0.007$ & $\textbf{0.89} \pm 0.006$ \\
& \textbf{768} & $0.88 \pm 0.004$ & $0.89 \pm 0.004$ & $0.90 \pm 0.004$ & $\textbf{0.92} \pm 0.005$ \\
\hdashline[.4pt/1pt]
\multirow{4}{*}{\textbf{TLDR}} 
& \textbf{128} & $0.51 \pm 0.008$ & $0.51 \pm 0.013$ & $0.52 \pm 0.012$ & $\textbf{0.53} \pm 0.006$ \\
& \textbf{256} & $0.55 \pm 0.01$ & $\textbf{0.57} \pm 0.01$ & $0.56 \pm 0.012$ & $0.55 \pm 0.01$ \\
& \textbf{384} & $0.55 \pm 0.009$ & $0.56 \pm 0.017$ & $0.58 \pm 0.01$ & $\textbf{0.59} \pm 0.007$ \\
& \textbf{512} & $0.56 \pm 0.012$ & $0.58 \pm 0.01$ & $\textbf{0.60} \pm 0.009$ & $0.59 \pm 0.01$ \\
\bottomrule
\end{tabular}
\end{table*}

\vspace{10mm}

\begin{figure*}[ht]
    \centering
    \begin{subfigure}[b]{0.48\textwidth}
        \centering
        \includegraphics[width=\textwidth]{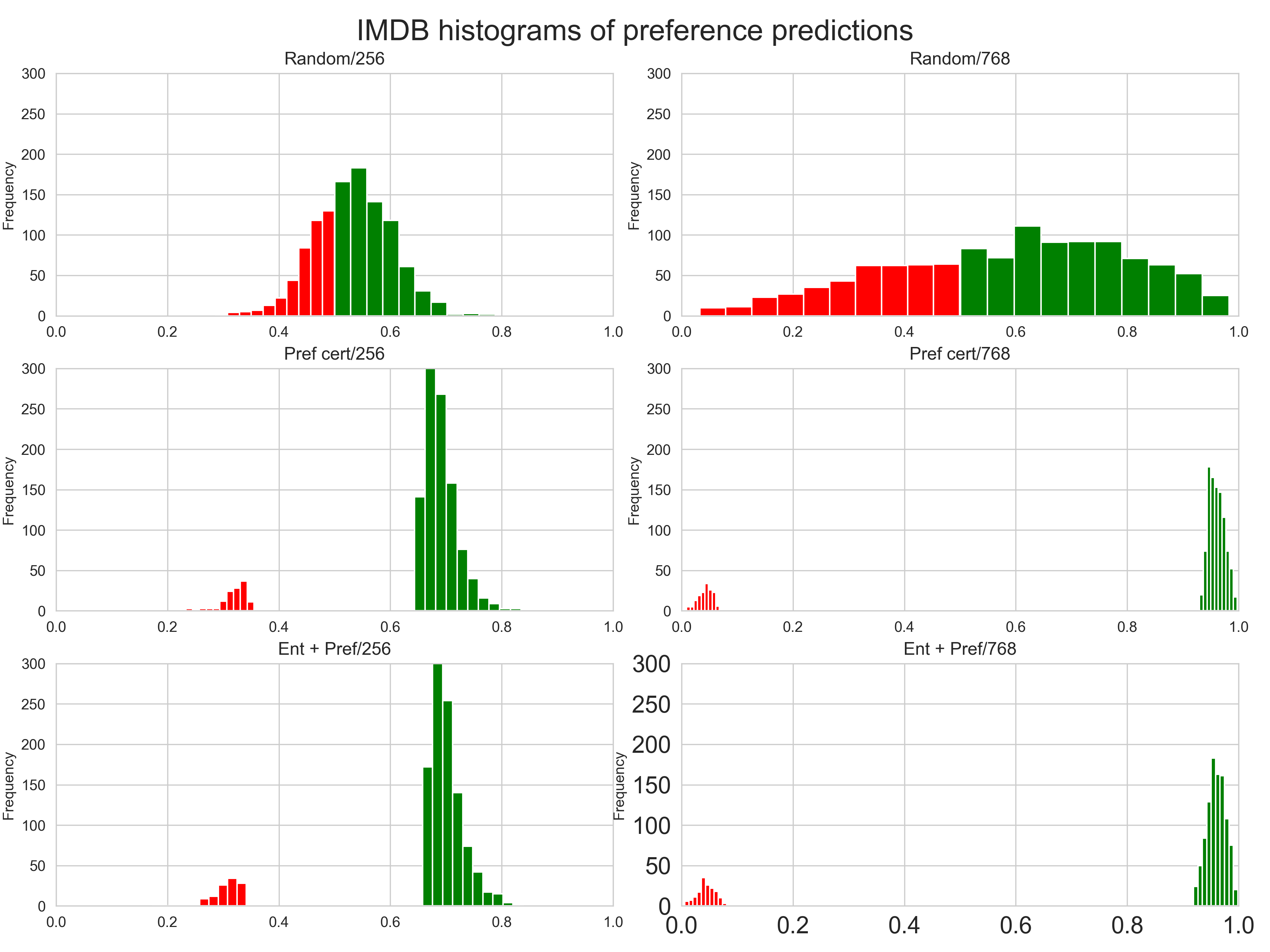}
        \label{fig:imdb_histogram}
    \end{subfigure}
    \begin{subfigure}[b]{0.48\textwidth}
        \centering
        \includegraphics[width=\textwidth]{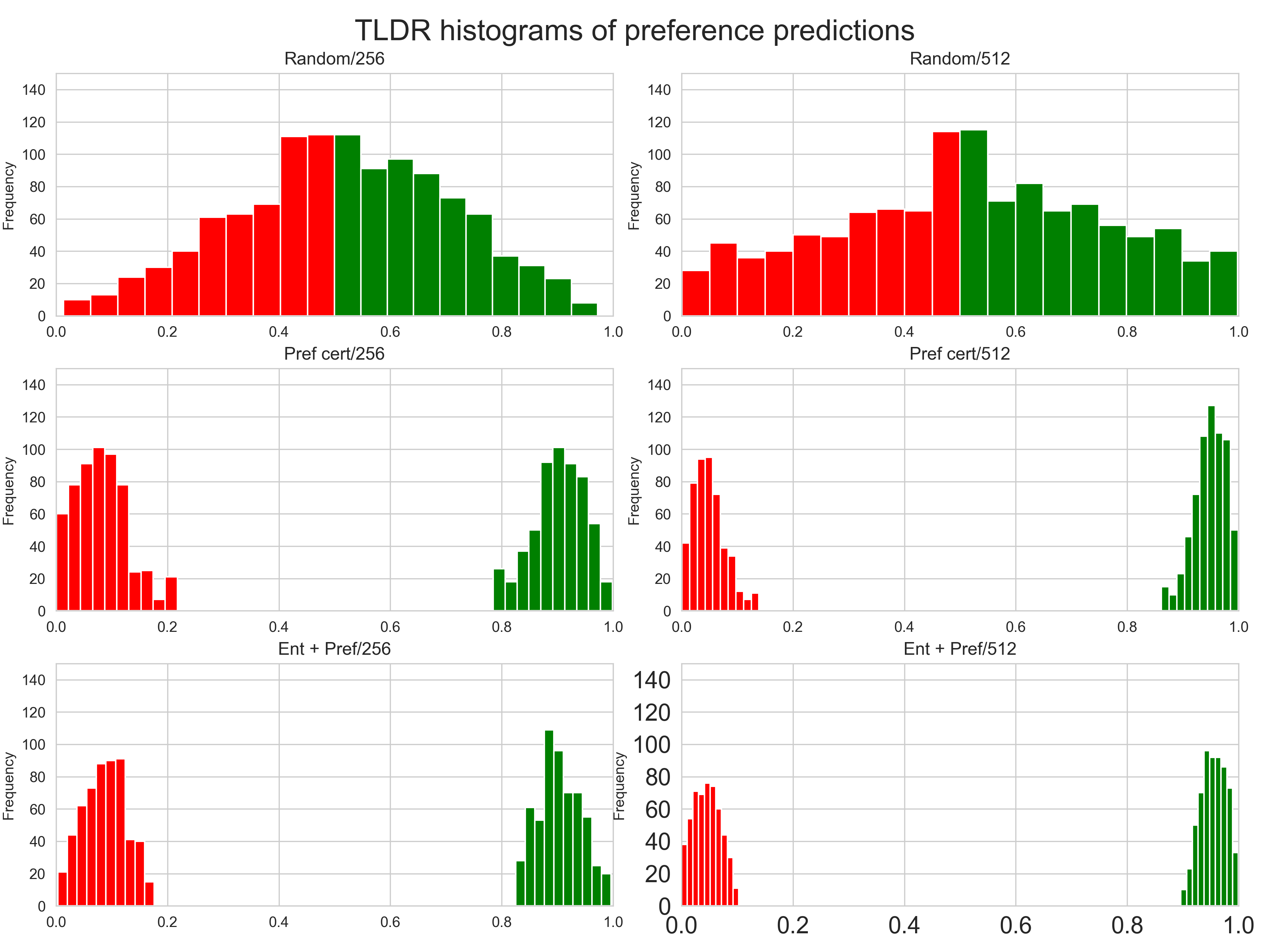}
        \label{fig:tldr_histogram}
    \end{subfigure}
    \caption{Histograms of probabilities from our implicit Bradley Terry preference model across a batch of acquired data; grouped by incorrect (red) and correct (green) preferences according to the oracle. This assumes a decision threshold of 0.5. Our preference certainty acquisition function surfaces confidently with wrong examples.}
    \label{fig:histograms}
\end{figure*}

\subsubsection{Analysing acquired data}
\label{sec:qualitative-results}

In section \ref{sec:acquisition_functions} we motivate why the preference certainty acquisition strategy may provide an advantage versus a random baseline when fine-tuning with DPO. This focused on whether it would surface examples where the implicit preference model provided an incorrect prediction, with certainty. We carry out a post hoc analysis of the data acquired during our experiments to better understand the characteristics of the acquired examples. In particular, what differs between the different acquisition strategies and how they change as fine-tuning phases progress. The approach we take is to look at how the implicit preference predictions from the model correlate with the true oracle preferences. 

We construct a classifier using the Bradley Terry (BT) model - equation 6 in \cite{rafailov2023direct} - that gives us $p(y_{1} \succ y_{0} |x) \in [0, 1]$ under our implicit reward model (equation \ref{eq:implicit_reward}). Using the probabilities provided, we construct histograms in figure \ref{fig:histograms} for the batches of $M$ acquired datapoints across all 9 seeds. We map the data in such a way that the bucket at 0.9 will contain examples where the BT model was most confidently correct according to it's probability, and 0.1 will contain the most confidently wrong. The red $0.0 \rightarrow 0.5$ contains all the incorrect predictions bucketed into 10 bins according to their probability. The green $0.5 \rightarrow 1$ contains all correct predictions. To determine correctness, we use a 0.5 decision threshold on our BT model and compare the result to the ranking provided by the oracle.

We can see from these histograms that the random acquisition selects quite uniform examples according to the implicit preference model predictions. The preference certainty-based acquisition on the other hand surfaces a lot of confidently incorrect examples which ultimately aid with improving fine-tuning performance when using DPO.

\section{Conclusion and Discussion}
\label{sec:conclusions}
We've demonstrated a simple and effective way to improve the use of an oracle labelling budget for preference fine-tuning LLMs. Our active learning setup builds upon DPO and uses the implicit preference model to determine which data points to get oracle judgements during online training.

Given the ever increasing computational cost involved in training SOTA large language models, it is important to consider the practical limitations of scaling up our setup. One such example is that we re-initialise the parameters of the model at each fine-tuning step $t$ as done in previous deep active learning works \cite{gal2017deep}. This helps us isolate the impact of the different acquisition strategies, which is the focus. A promising direction of future work is to integrate approaches from online learning \cite{ritter2018online}. This could significantly improve computational efficiency by allowing us to not re-initialise the parameters at each time step and spend the majority of the fine-tuning budget on the most recently acquired data. This could involve further changes to the model and/or how we are sampling the data when fine-tuning. In Appendix \ref{app:online}, we discuss minimally adapting our approach for online learning and present promising preliminary results to motivate future work in this direction. 

% I suggest using this: An alternative future direction is to explore combining our approach with parameter-efficient fine-tuning techniques, such as LORA~\cite{hu2021lora}. Implementing smaller batches with more regular updates could potentially enhance the effectiveness of our active approach. Additionally, exploring further data acquisition strategies could prove fruitful. This might include measures to increase the diversity of samples within a batch or adopting a more Bayesian approach to explicitly model the epistemic uncertainty of our model \cite{kirsch2019batchbald}. Furthermore, the use of LLMs as evaluators in this context presents an intriguing possibility. Investing more effort into developing smaller, more economical models for such use cases would facilitate conducting extensive ablations, thereby allowing for more robust conclusions. We leave these as promising future directions.

An alternative direction here is to explore combining our approach with parameter-efficient fine-tuning techniques like LORA \cite{hu2021lora}. Acquiring smaller batches with more regular updates would also likely further favour the more active approach. Another interesting direction of future work is to explore additional data acquisition strategies. For example, we can include measures of the diversity of samples within a batch, or take a more Bayesian approach to explicitly model the epistemic uncertainty of our model \cite{kirsch2019batchbald}. Lastly, the use of LLMs as evaluators in this setting is of independent interest. Investing more time into getting smaller, more economical models to work for these sorts of use cases would make it easier to run larger amounts of ablations in order to draw stronger conclusions.

\section*{Impact Statement}
We deal with the problem of fine-tuning large language models. Although the models used in our specific experiments can fit on a single large A100 GPU and are manageable in terms of energy consumption, our framework could be applied to much larger closed-source models. This could lead to the indirect negative consequence of this work on the environment, due to the large amount of energy required.

On the positive side, we focus on the problem of how to better use Human and AI feedback to align large language models as part of a fine-tuning process. This could have a positive impact on AI safety research.

\bibliography{main}
\bibliographystyle{icml2024}

\newpage
\appendix
\onecolumn

\section{Oracle prompts}
\label{app:oracle-prompts}
\begin{figure}[h]
\vspace{5mm}
\centering
\begin{minipage}[t]{.45\textwidth}
\tiny % Set the font size to tiny for the first minipage
\begin{verbatim}
// SENTIMENT ORACLE PROMPT 

<SYSTEM>
You are a helpful assistant that evaluates the quality and 
positive sentiment of movie reviews
</SYSTEM>

<USER>
Which of the following movie reviews is better? The best one 
will be the one with the most positive sentiment, which also
is grammatically correct, consistent, and avoids repetition.

Review A:
{{PROMPT}} {{COMPLETION-A}}

Review B:
{{PROMPT}} {{COMPLETION-B}}

First, provide a one-sentence comparison of the two reviews, 
explaining which is better and why. Second, on a new line, 
state only "A" or "B" to indicate your choice. 

You must choose A or B for the preferred answer even if 
neither review is very good.  

Your response should use the format:
Comparison: <one-sentence comparison and explanation>
Preferred: <"A" or "B"> 
<\USER>
\end{verbatim}
\end{minipage}\hfill
\begin{minipage}[t]{.45\textwidth}
\tiny % Set the font size to tiny for the second minipage
\begin{verbatim}
// SUMMARIZATION ORACLE PROMPT 

<SYSTEM>
You are a helpful assistant that evaluates the quality of 
summaries for internet posts.
</SYSTEM>

<USER>
Which of the following summaries does a better job of 
summarizing the most important points in the given 
forum post, without including unimportant or irrelevant 
details?

Post:
{PROMPT}

Summary A:
{COMPLETION_A}

Summary B:
{COMPLETION_B}

First, provide a one-sentence comparison of the two 
summaries, which you prefer and why. Second, on a new line, 
state only "A" or "B" to indicate your choice. 

You must choose A or B for the preferred answer even 
if neither summary is very good.

Your response should use the format:

Comparison: <one-sentence comparison and explanation>
Preferred: <"A" or "B">
<\USER>
\end{verbatim}
\end{minipage}
\caption{GPT-4 oracle prompts for sentiment and summarization tasks.}
\label{fig:gpt-4-oracle-prompts}
\end{figure}

\section{Data preprocessing}
\label{app:data-pre-processing}
For IMDB, each sample $x$ is randomly drawn beginning of a review. The only processing we do here is to randomly truncate $x$ to a number of tokens randomly drawn from the range 8-16 tokens. See table \ref{tab:imdb-data-examples} for some truncated examples that we feed to the model to complete a positive review for:

\begin{table}[h]
    \centering
    \small
    \begin{tabular}{|l|}
    \hline
    \textbf{Truncated movie review prompt samples} \\
    \hline
        I very much looked forward to this movie. Its a good family ... \\
        Really, I can't believe that I spent \$5 on this movie. I am a huge zombie  ... \\
        I have read all of the Love Come Softly books.... \\        
        I've seen all four of the movies in this series. Each one strays further ... \\        
    \hline
    \end{tabular}    
    \vspace{4mm}
    \caption[IMDB movie review samples]{IMDB data from \url{https://huggingface.co/datasets/imdb}; randomly truncated to produce a prompt for training data generation and evaluation.}
    \label{tab:imdb-data-examples}
\end{table}

For TLDR, we filtered the Reddit posts between 200-1000 characters. This was mainly due to memory contraints of the GPUs used to train the models. We also filtered whole broad categories of Reddit posts out, such as r/offmychest and r/tifu, because they had high likelihood of containing explicit content. Finally we removed trailing space tokens. See table \ref{tab:tldr-data-examples} for examples.

\begin{table}[h]
    \centering
    \scriptsize
    \begin{tabular}{lp{11.5cm}}
        \toprule
        \textbf{Prompt} & SUBREDDIT: r/cats TITLE: Acquired cat! Now a question.. POST: So, I just got a lovely little cat named Luna. She's about a year, a year and a half and pretty tiny. I live in an apartment located on the 5th floor of my building. My apartment doesn't have AC (I'm in NYC) and I usually like to leave the windows open for ventilation. They've got child bars, but because Luna is so small she can easily fit through them--and did a few moments ago. Nearly gave me a heart attack watching her slip through them and walked out onto a very narrow ledge 5 floors above a concrete sidewalk. She came right back in, but now I'm concerned about having a dead cat on my hands (or more accurately, on my sidewalk). So my question is, should I trust her cat instincts and leave the windows open? Or shall I sit in a stuffy apartment with the windows sealed? TL;DR: \\
        \midrule
        \textbf{Human Summary} & I live on the 5th floor and my cat just walked out on my window ledge and came back in. Should I be nervous she's going to explore too far out and fall to her kitty death?\\
        \toprule
        \textbf{Prompt} & SUBREDDIT: r/AskReddit TITLE: I want to throw a great party, Reddit. What are some special things that I can do to make this happen? POST: Alright, so my 19th birthday is next Friday. I live in a college town, renting a house alone for the summer, and I want to throw a great party. I have invited friends from both school (which means they are coming from all over the state) and from home. However, since it's summer, I know that a lot of people won't show. Gas is expensive and for some, it's a far drive, so I understand. I'm thinking music, card playing, beer pong, etc... But what else can I do to make sure that everyone mingles, it isn't lame, and that things stay under control while everyone still has a good time? TL;DR:\\
        \midrule
        \textbf{Human Summary} & How do I throw a kick ass party with a bunch of people who don't really know each other and my house is still standing in the morning?\\
        \bottomrule        
    \end{tabular}    
    \vspace{4mm}
    \caption[TLDR Reddit post samples]{Samples of TLDR data from \url{https://huggingface.co/datasets/CarperAI/openai_summarize_tldr}}
    \label{tab:tldr-data-examples}
\end{table}

\section{Example responses}
\label{app:apo-samples}
In tables \ref{tab:imdb_oracle_responses}, \ref{tab:tldr_oracle_responses} we provide examples of prompt completion pairs for our models discussed in section \ref{apl:exp-models} after the fine-tuning process alongside the preferred GPT-4 oracle choice. 

\begin{table}[h]
    \centering
    \scriptsize
    \begin{tabular}{lp{8.5cm}}
        \toprule
        \textbf{Prompt} & The comments of the previous user are harsh indeed \\
        \textbf{A: Fine-tuned completion} & . This is a great film with a great story and a great cast. Highly recommended. I saw this movie on Sund \\
        \textbf{B: Original completion} & . I am a fan of the original series and I have watched the first two seasons of the series. I have also \\
        \textbf{Preference} & A \\
        \textbf{Rationale} & Review A is better as it expresses a clear positive sentiment towards the film, while Review B does not provide any opinion on the movie \\
        \midrule
    
        \textbf{Prompt} & I was looking forward to this so much, being a \\
        \textbf{Fine-tuned completion} & huge Harry Potter fan. I was very impressed with the story and the acting. I highly recommend it.\\
        \textbf{Original completion} & huge fan of the original series. I was so disappointed. I was expecting a lot more from this. \\
        \textbf{Preference} & A \\
        \textbf{Rationale} & Review A is better because it provides a positive sentiment about the movie, is grammatically correct, and does not contain repetition like Review B. \\
        \midrule
        \midrule
    
    \end{tabular}    
    \vspace{4mm}
    \caption[Completion samples from fine-tuned model IDMB]{Sampled completions, and GPT-4 oracle evaluation from an IMDB fine-tuned with certainty acquisition.}
    \label{tab:imdb_oracle_responses}
\end{table}

\begin{table}[ht]
    \centering
    \scriptsize
    \begin{tabular}{lp{8.5cm}}
        \toprule
            \textbf{Prompt} & SUBREDDIT: r/relationships
            TITLE: I(17M)am gealous because my girlfriend(16F) of over 4 months is having conversation that she starts with my classmate.
            POST: I am now really down because of this. We had an argument 2 months ago about this and I told her to stay away from my classmate(he is a really bad person and always speaks really bad about her when he is with others or even with me, we've been colleagues for 10 years now and he is awful, I hate him), and I'm sick of it. 
            The problem is, today after checking her phone I realized she starts most conversation and he is mostly not answering to her immediately. 
            I am not sure what to do and I am really confused and really really sad and disappointed. What should I do?
            TL;DR:\\
            \textbf{A: Fine-tuned completion} &  
            
            GF of over 4 months starts most conversation with my classmate, he is mostly not answering to her immediately. I am not sure what to do and am really sad and disappointed. Need advice. [endoftext]\\
            \textbf{B: Original completion} &  
            
            I am really sad because my girlfriend starts most conversation with my classmate and I'm sick of it. What should I do?[endoftext]\\
            \textbf{Preference} & A\\
            \textbf{Rationale} & Summary A is preferred as it includes more relevant details from the post, such as the duration of the relationship, the classmate's lack of immediate response, and the poster's need for advice, which were missing in Summary B.\\
            \midrule
            \textbf{Prompt} & SUBREDDIT: r/dogs
            TITLE: Need a little advice regarding our elderly dog.
            POST: Our rescue [Lt. Ellen Ripley] has been showing her age lately. Actually for the last six months or so she has been going a bit downhill. Formerly very active and happy to play she's blind, deaf and now sleeps or spins in circles all day. She hasn't been able to hold her water or poops at all and we've been coming home to a nightly cleanup. Worse she's been biting at us a lot and seems to think our hands are actually delicious fish sticks. 
            
            We've tried medicine and have seen the vet numerous times but she just seems to be getting worse and worse.
            
            So, advice time. Am I being selfish? I feel like keeping her around may not be the best thing for her anymore. I worry that she spends her time now scared, confused and in the dark. But so far as I know, she's not in any pain and she can still recognize me on occasion.
            
            I'll
            TL;DR:\\
            \textbf{A: Fine-tuned completion} &  
            
            Old rescue dog has been getting worse and worse. Tried medicine and the vet. Don't know if keeping her around is the best thing for her anymore. Need advice.[endoftext]\\
            
            \textbf{B: Original completion} &  
            
            We have a dog that is getting older and is starting to become more depressed. What should we do about it?[endoftext]\\
            \textbf{Preference} & A\\
            \textbf{Rationale} & Summary A is more accurate and detailed, capturing the severity of the dog's condition, the attempts at treatment, and the owner's dilemma, while Summary B oversimplifies the situation by referring to the dog as merely "more depressed."\\
            \midrule
    
    \end{tabular}    
    
    \caption[Completion samples from fine-tuned model TLDR]{Sampled completions, and GPT-4 oracle evaluation from an TLDR fine-tuned with certainty acquisition.}
    \label{tab:tldr_oracle_responses}
\end{table}

\section{Online variation}
\label{app:online}
In section \ref{sec:conclusions} we note that a practical limitation to scaling algorithm \ref{alg:APO-training} to significantly larger models is that the parameters $\theta_{t}$ are reset to $\theta_{0}$ in step 12 and the model is fine-tuned on all the data acquired so far each step. Here we motivate an online variant where we instead don't reset the parameters and do a single gradient update with respect to the most recently acquired data during fine-tuning. This significantly reduces the time spent fine-tuning during the active learning approach. This is similar to the approach taken in \cite{guo2024direct}, where they randomly, instead of actively, acquire the data and consider comparing online vs offline.

\begin{figure}[h]
    \centering
    \includegraphics[width=0.5\textwidth]{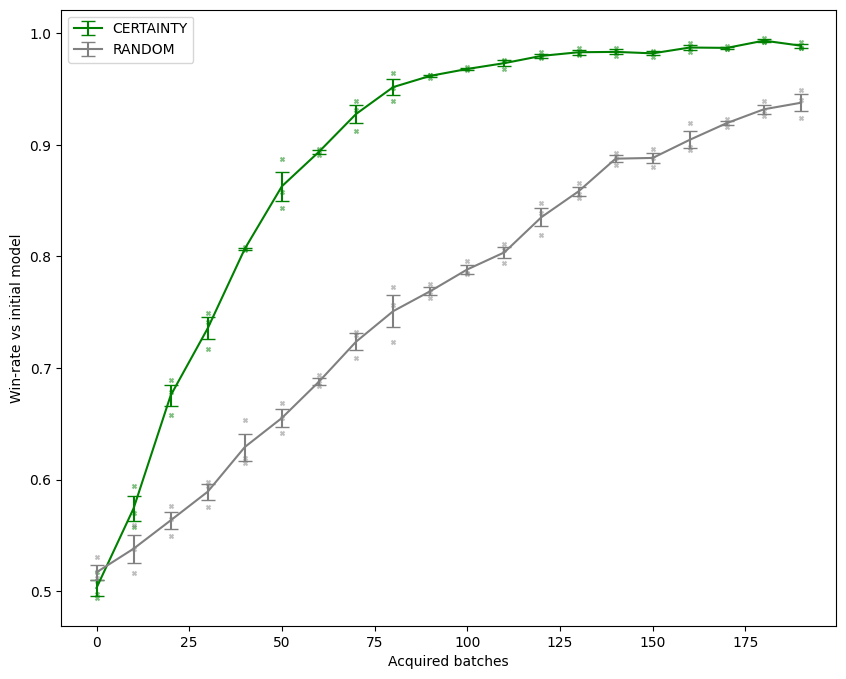}
    \caption[Provisional online fine-tuning results on IMDB]{Win-rate vs initial model after each acquired batch for IMDB with random and preference certainty acquisition and online fine-tuning. Only a single fine-tuning gradient step is taken on the latest batch.}
    \label{fig:online-learning}
\end{figure}

We re-run our IMDB experiment from section \ref{sec:apl_experiments} with this online active variation for random and preference certainty acquisitions. We do 3 random seeds and include standard error bars in figure \ref{fig:online-learning}. We find that preference certainty significantly outperforms random on this problem, motivating further study.

\clearpage
\section{Fine-tuning iterations}
\label{app:apl-convergence}
\begin{figure}[h]
     \centering
     \begin{subfigure}[b]{0.45\textwidth}
         \centering
        \includegraphics[width=\textwidth]{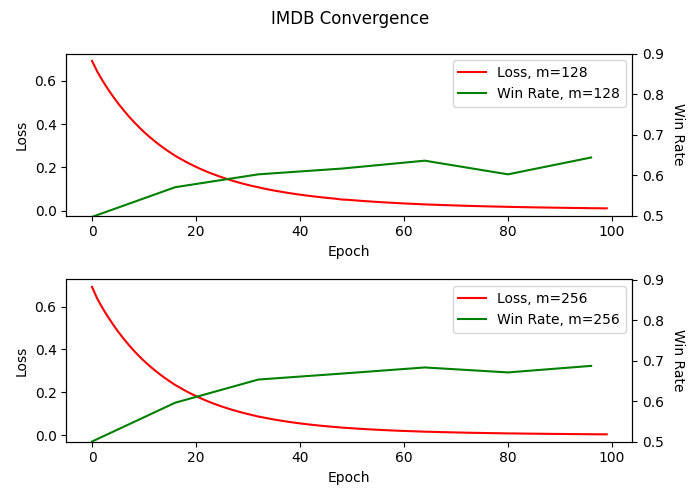}
     \end{subfigure}
     \begin{subfigure}[b]{0.45\textwidth}
         \centering
        \includegraphics[width=\textwidth]{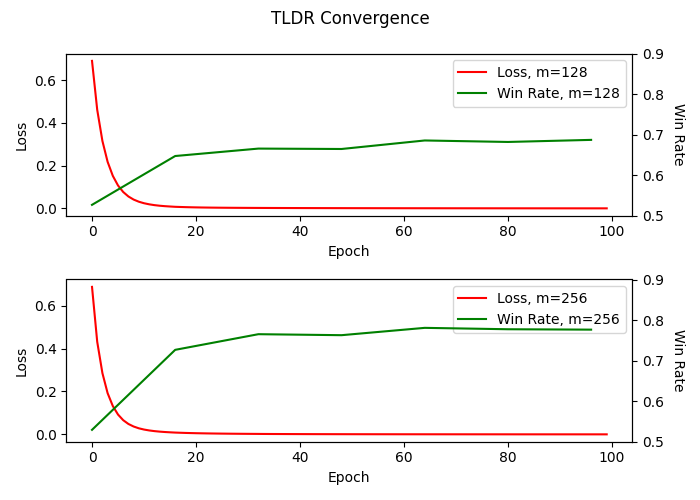}
     \end{subfigure}
     \caption[Convergence behaviour of fine-tuning models]{Illustrates a sample of how the convergence of the loss relates to the win-rate. Used for empirically inferring the number of fine-tuning epochs to apply after each data acquisition step.}
     \label{fig:apl-convergence}
\end{figure}

In order to determine how many fine-tuning epochs to carry out after each new data acquisition step, we took an empirical approach of defining a fixed number of epochs. We on the number of epochs it took on average for the model to converge at different dataset sizes. We analysed loss and win-rate curves (on a hold out validation set) for the different model and dataset combinations and decided upon 50 epochs for IMDB and 70 for TLDR - see figure \ref{fig:apl-convergence} for a sample of convergence behaviour.

\end{document}